# An HPSG Parser Based on Description Logics*

**J. Joachim Quantz**

Technische Universität Berlin, Projekt KIT-VM11, FR 5–12,
Franklinstr. 28/29, D-10587 Berlin, Germany, e-mail: jjq@cs.tu-berlin.de

## Abstract

In this paper I present a parser based on Description Logics (DL) for a German HPSG-style fragment. The specified parser relies mainly on the inferential capabilities of the underlying DL system. Given a preferential default extension for DL disambiguation is achieved by choosing the parse containing a qualitatively minimal number of exceptions.

## 1 Introduction

In this paper I present a parser for HPSG based on Description Logics (DL). The main motivation for specifying such a parser relies on considerations concerning the disambiguation of NL expressions. In [Schmitz, Quantz 93] it is shown how different types of ambiguity can be handled with a homogeneous approach based on the notion of preference rules [Jackendoff 83]. A major requirement for such a unified approach is that information usually represented rather differently (e.g. syntactic, semantic, and encyclopedic information) is homogeneously represented in a uniform and declarative formalism in order to express and evaluate the complex preferences stemming from the different kinds of information.

Description Logics have been developed in the field of Knowledge Representation (see, for example, [Brachman et al. 91]). They have already been used for the representation of semantic and encyclopedic information, e.g. [Allgayer et al. 89, Stock 91, Preuss et al. 92]. Due to their similarity to typed feature formalisms [Carpenter 92], syntactic information is in principle also expressible in DL, as already sketched in [Quantz 93, Quantz, Schmitz 94]. Furthermore, Preferential Default Description Logics (PDDL) based on weighted defaults [Quantz, Ryan 93] can be used to represent the preference rules in a declarative and formally well-founded way.

In the following I will mainly show how HPSG-style syntactic information can be represented in DL, and how a simple parser can be build by using the inference capabilities of a DL system. Note that when specifying the parser I will keep the presentation as simple as possible, thereby deliberately ignoring efficiency aspects. I will also refrain from modeling *all* aspects of relevant knowledge in DL, i.e. there are still pieces of information which are not explicitly encoded in the DL modeling, but are rather implicitly contained in the parser (e.g. information related to linear precedence and to traces).

The main objective of this paper it thus neither to contribute to research in efficient parsing technology, nor to develop a declarative formalism in which *all* aspects relevant for NLP can be represented. It is rather to provide the basis for an implementation of the exception minimization approach to interpretation proposed in [Quantz 93]. In Section 6 I will briefly sketch how the DL-based parser presented in Section 5 can be extended to realize this approach.

## 2 Basic Ideas

An important distinction made in DL, but missing in traditional feature formalisms, is the one between *objects* and *types*. DL formulae either express that a type $t_1$ is more specific than (or subsumed by) a type $t_2$ ($t_1 :< t_2$) or that an object o is an instance of a type or, using DL terminology, a *concept* (o :: c).

Applying this schema to the task of NLP, we can say that the objects in this domain are words or phrases, and that the types are syntactic categories. Furthermore, given a phrase $o_1$ we have additional relations between this phrase and its constituents $o_2, o_3, \ldots$, usually expressed as "$o_2$ is a daughter of $o_1$". In DL this is modeled as '$\langle o_1, o_2 \rangle$ :: dtrs', or equivalently as '$o_1 :: dtrs{:}o_2$'. 'dtrs' thus acts as a binary predicate or, using DL terminology, as a *role*. Note that roles can have more than one value in contrast to features, which are functional. We can thus write '$o_1 :: dtrs{:}o_2 \ \& \ dtrs{:}o_3$'.

Note further that the objects stand for occurrences of words or phrases, and that different occurrences of the same word will be represented by different objects. This is represented by writing '$o_2 :: phon{:}er$', for example, to express that $o_2$ is an occurrence of the form 'er'.

This is all rather similar to standard HPSG notation, and the main difference is that in addition to the feature structures used in HPSG, we add an additional level of objects, which we see as instances of the feature structures. Feature structures thus correspond to types or more precisely to DL concepts. In a way, the objects in DL are used to make the HPSG feature structures *persistent*, i.e. to have pointers or names to refer to them.

The additional level of objects allows a straightforward description of the parsing task. We start with a number of objects, namely words, whose phonological value and position is known. We want to end up with a single object containing all these words as (not necessarily immediate) constituents. Now the immediate dominance schemata in an HPSG tell us how to construct phrases from words or other phrases. Thus the main operation for building a phrase is to create a new object being an instance of an ID schema (note that ID schemata are feature structures and therefore concepts) and to fill in the required daughters by using the objects available as building material. This is achieved by choosing the 'functor-daughter' and filling the required arguments.

*The project KIT-VM11 is supported by the German Federal Minister of Research and Technology (BMFT) under contract 01 IV 101Q8.

Three points are important in the following sections:

1. Obviously, objects cannot be combined in a random way. In HPSG the ID schemata and the lexical entries contain information concerning combination with other phrases. I will model this information in DL and use standard DL inferences to check consistency of combinations. Thus the DL system is used to perform the unification task underlying HPSG and similar Unification Grammars.

2. An object can only be used as building material for a phrase if it has not yet been used as building material for some other phrase. Furthermore, when looking for daughters of a new phrase, we have to fill those daughters for which a filler is required, but not yet specified. I will use the *epistemic operator* **k** proposed in [Donini et al. 92] to formalize these intuitions and then use standard DL retrieval for checking these constraints.

3. For syntactically ambiguous expressions there is more than one possibility to combine the words/phrases. Since the objects and especially the relations between them are viewed from different perspectives in the alternative interpretations, we need a mechanism in DL to represent these different views. I will use *situated descriptions* 'o :: c in s' in the following to formalize this notion of different perspectives. There is a rough correspondence between the situations used to capture the specific interpretations and the charts created in chart parsing.

## 3 The Underlying Description Logic

Description Logics vary wrt the term-building operators they contain. In this section I will present the syntax of the DL which is used in the examples given in the next two sections. Due to space limitations I will not specify the formal semantics for this DL (see, for example, [Hoppe et al. 93, Quantz, Schmitz 94] for a model-theoretic semantics):

$$
\begin{aligned}
t &\rightarrow c, r, t_p, t_1 \& t_2, \mathbf{k}(t) \\
c &\rightarrow \text{the}(r,c), \text{some}(r), \text{no}(r), \text{exactly}(n,r) \\
  &\quad r{:}o, r_1 = r_2 \\
r &\rightarrow \text{feat}, \text{domain}(c), \text{range}(c), r_1.r_2, \text{inv}(r) \\
\gamma &\rightarrow t_1 :< t_2, t_1 := t_2, c_1 => c_2 \\
  &\quad o :: c \text{ in } s, \text{extend\_sit}(s_1, s_2)
\end{aligned}
$$

When specifying the fragment and the parser in the next sections I will use a notation based on the PROLOG interface provided by the BACK system [Hoppe et al. 93]. In BACK a distinction is made between term introductions or definitions, and constraint-like rules. A term name can be introduced either as *primitive* ($t_n :< t$), i.e. only necessary conditions are given, or as defined ($t_n := t$), i.e. necessary and sufficient conditions are given. A rule $c_1 => c_2$ means that each object being an instance of $c_1$ is also an instance of $c_2$.

The formula 'extend\_sit($s_1,s_2$)' expresses the fact that situation $s_2$ is an extension of situation $s_1$. This means that 'o :: c in $s_1$' implies 'o :: c in $s_2$' for all objects o and concepts c.

In order to distinguish between telling and querying information I will use 'o :: c in s' for tells and 'o ?: in s' for queries. I furthermore assume that a tell only suceeds if it is consistent with the previously entered information; otherwise it fails. When the object used in a query is a variable, the system will retrieve all known instances of a concept, i.e. 'Object ?: in s' will return the objects known to be instances of 'c' in 's' by backtracking.

Note that the epistemic operator **k** will only be used in queries. It can therefore be straightforwardly integrated into existing DL systems. Since this is also true for situated descriptions, the parser presented in Section 5 is largely based on standard inference capabilities of DL systems.

## 4 A Small Fragment

In this section I will present examples from an HPSG-style fragment for German modeled in DL. Due to space limitations I will not specify all the information contained in this modeling but only the one needed to illustrate the main characteristics of the formalization and the example sentence 'Die schöne Frau sieht sie' discussed in the next section.

The fragment is based on the presentation in [Pollard, Sag 87] and its application to German in [Hiltl 91]. A main difference between my DL modeling and standard HPSG modeling is that I avoid feature pathes which would introduce superfluous DL objects. There is thus no feature 'head' in my modeling since it would yield the introduction of head objects whose ontological status seems controversial. Consequently, my Head Feature Principle specifies equivalence not for a single feature 'head', but rather for each head feature separately.

The fragment contains five main categories, namely *noun*, *np*, *verb*, *det*, and *adj*. For illustration, the definitions of *noun* and *np* are given below:

```
noun  :=  maj:n & lex:+
np    :=  maj:n & lex:–
```

Phrase structure is represented by roles as the following:

```
     dtrs  :<  domain(sign) & range(sign)
comp_dtrs  :<  dtrs
comp_dtr1  :<  comp_dtrs & feat
 head_dtr  :<  dtrs & feat
functor_dtr :< dtrs & feat
```

The feature 'functor\_dtr' will be used by the parser to specify the sign acting as functor of a new phrase. Its value will be identical to the value of 'head\_dtr', 'adj\_dtr', or 'filler\_dtr', depending on the particular *Immediate Dominance* (ID) schema used. Note that the daughters which are modeled as features are functional, i.e. no phrase can have two fillers for 'head\_dtr'.

Corresponding to these daughter roles and features we have argument roles and features as 'comp\_arg1' etc. I then distinguish the following types of phrase structures:

```
comp_structure   :=  some(head_dtr) &
                     functor_dtr=head_dtr &
                     no(adj_dtr) & no(filler_dtr)
 adj_structure   :=  some(adj_dtr) & some(head_dtr) &
                     functor_dtr=adj_dtr &
                     no(comp_dtrs) & no(filler_dtr)
filler_structure :=  some(filler_dtr) & some(head_dtr) &
                     functor_dtr=filler_dtr &
                     no(comp_dtrs) & no(adj_dtr)
```

Thus in a 'comp\_structure' the 'head\_dtr' acts as a functor. Note that it has to be explicitly stated whether a certain feature is empty, e.g. 'no(adj\_dtr)' for 'comp\_structure'. DL systems assume an open world and take all descriptions as being partial, i.e. the fact that there is currently no known filler for a role at an object does not imply that there will never be one.

The fragment contains six ID schemata, namely three for noun phrases, one for verb phrases, one for adjuncts, and one for topicalization.

```
id1  :=  comp_structure &
         the(head_dtr,np & nform:comm) &
         some(comp_dtr1) & no(comp_args)
id2  :=  comp_structure &
         the(head_dtr,noun & nform:comm) &
         no(comp_dtrs) & some(comp_arg1)
id3  :=  comp_structure & the(head_dtr,verb) &
         no(comp_args) & mc:–
id4  :=  adj_structure & the(adj_dtr,adjunct)
id5  :=  filler_structure &
         the(head_dtr,maj:v & no(comp_args)) &
         mc:+ & the(filler_dtr,top:+)
id6  :=  comp_structure &
         the(head_dtr,noun & nform:pro) &
         no(comp_dtrs) & no(comp_args)
```

For the lexical entries I will use three morpho-sytntactic features (*nform*, *case*, *gen*) to illustrate agreement between nouns, adjectives, and determiners. Agreement concerning case and gender between nouns and determiners is modeled by specifying that the value of the feature 'case' at a common noun is the same as the value of the feature 'case' at the object filling the feature 'comp_arg1' (which is the determiner).

Below are lexical entries for 'frau' and 'sie':

```
noun & nform:comm  =>  exactly(1,comp_args) &
                       the(comp_arg1,det) &
                       case=comp_arg1.case &
                       gen=comp_arg1.gen)
noun & nform:pro   =>  no(comp_args)
lexeme:frau        =>  noun & nform:comm & gen:f
phon:frau          =>  lexeme:frau
lexeme:er_sie      =>  noun & nform:pro
phon:sie           =>  lexeme:er_sie & gen:f
```

Note the hierarchical nature of the modeling—the subcategorization information is specified for common nouns and pronouns in general, and is then inherited by each specific common noun and pronoun. Information shared by all forms of a lexeme is specified as a property of the lexeme, whereas information specific to a particular form is specified for this form only.

Adjectives require non-saturated noun phrases as arguments and agree with them wrt case and gender:

```
adj            =>  adjunct & case=mod_arg.case &
                   gen=mod_arg.gen &
                   the(mod_arg,np & some(comp_args))
lexeme:schoen  =>  adj
phon:schoene   =>  lexeme:schoen
```

Finally, the lexical entry for 'sieht':

```
verb          =>  the(comp_arg1,np & case:nom)
lexeme:sehen  =>  verb & exactly(2,comp_args)
                  the(comp_arg2,np & case:acc)
phon:sieht    =>  lexeme:sehen
```

Note that for verbs taking more than two arguments we need additional features 'comp_arg3' and 'comp_arg4'.

In addition to the information modeled so far we need a formalization of the principles underlying the combination of signs in HPSG. Some of these principles hold only for phrases and not for signs in general. A phrase is defined as follows:

```
phrase  :=  some(dtrs)
phrase  =>  lex:–
lex:–   =>  phrase
```

The *Head Feature Principle* is then defined as:

```
phrase  =>  maj=head_dtr.maj &
            gen=head_dtr.gen &
            case=head_dtr.case
```

The parsing process presented in the next section is essentially triggered by signs which can act as functors, namely signs with unsaturated subcat lists, signs with slashes, and pronouns:

```
some(args)        =>  functor
some(slash)       =>  functor
noun & nform:pro  =>  functor
```

## 5  DL-Based Parsing

In this section I will present the basic structure of a DL-based parser for the above fragment. The parser is realized by five main predicates. I assume that the initial information given to the parser consists of descriptions of the words occurring in the expression to be parsed. Consider the ambiguous sentence

(1) Die schöne Frau sieht sie.

(2) The pretty woman sees her.

(3) The pretty woman she sees.

The initial DL representation of this sentence is:

$w_1$ :: phon:die & start:0 & end:1 in $s_1$
... :: ...
$w_5$ :: phon:sie & start:4 & end:5 in $s_1$

Given this information the parser builds phrases from the five words. This is done by creating new phrases until no more combinations of signs are possible. The parsing succeeds if the words have been all used up and a single phrase results:

```
parse_sign(Sit,Sit) :-
    findall(Sign,Sign ?: sign & no(k(inv(dtrs))) in Sit,[_]).
parse_sign(Sit,FinSit) :-
    new_phrase(Sit,NewSit),
    parse_sign(NewSit,FinSit).
```

Note that the epistemic concept 'no(k(inv(dtrs)))' is used to determine whether a sign is still available for phrase building. An object is an instance of this concept if it is not a filler of some 'dtrs' role at any other object.

The basic idea of building a new phrase is to look for a sign which can act as a functor, to choose an ID schema in which this sign is a functor, and to find the required arguments of the functor. Finally, the linear precedence rules are checked and, if necessary, traces are introduced.[1]

```
new_phrase(Sit,FinSit) :-
    Sign ?: functor & no(k(inv(dtrs))) in Sit,
    select_id_schema(Sign,Sit,Phrase,NewSit),
    complete_arguments(Sign,NewSit,NextSit),
    check_lps_and_continuity(Phrase,Sit,NextSit,FinSit).
```

Selection of an ID schema is realized in a rather naive and simple way—we just take an ID schema and try to create a new phrase as an instance of this schema, where the feature 'functor_dtr' is filled by the functor.

---

[1]Due to space limitations I do not specify the predicate 'check_lps_and_continuity' in this paper.

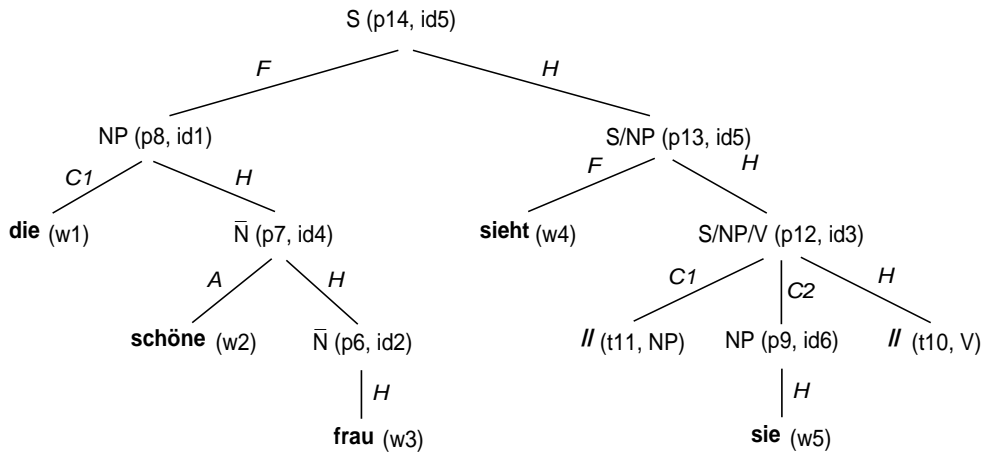

Figure 1: Phrase structure of the sentence 'Die schöne Frau sieht sie' (The pretty woman sees her). 'H' stands for 'head_dtr', 'C' for 'comp_dtr', 'A' for 'adj_dtr', and 'F' for 'filler_dtr'.

```
select_id_schema(Sign,Sit,Phrase,NewSit) :-
    id_schema(ID),
    extend_sit(Sit,NewSit),
    Phrase :: ID & functor_dtr:Sign in NewSit.
```

Information about existing ID schemata thus has to be encoded as facts of the form 'id_schema(id1)', etc. The predicate 'extend_sit(Sit,NewSit)' is used to tell the DL system to create a new situation which is an extension of the current situation.

Note that no further knowledge about the actual modeling of ID schemata is used in the parser except for the fact that each ID schema has a 'functor_dtr'. Note further that the DL tell will fail if the information known about the functor cannot be unified with the information required by the ID schema for the filler of 'functor_dtr'.

In order to complete the arguments of the functor, the parser checks for each argument feature ArgFeat whether an argument is required (some(ArgFeat)) but not yet specified (no(k(ArgFeat))). If so, 'find_arg' looks for such an argument and enters it as filler for ArgFeat. Then the remaining arguments are completed.

```
complete_arguments(Functor,Sit,FinSit) :-
    arg_feature(ArgFeat),
    Functor ?: some(ArgFeat) & no(k(ArgFeat)) in Sit,
    !,find_arg(Functor, Sit,ArgFeat,NewSit),
    complete_arguments(Functor,NewSit,FinSit).
complete_arguments(_,Sit,Sit).
```

Again we need to introduce facts specifying the arguments used in the fragment, e.g. 'arg_feature(comp_arg1)'.

If an argument is required it has to be filled, therefore the Cut. Thus the recursion terminates successfully only when all required arguments are actually filled. Note that the only information about argument structure needed by the parser are facts of the form 'arg_feature(comp_arg1)' for all argument features.

To find an argument the parser looks for a sign which has not yet been used for phrase building and asserts it as filler for the argument feature. Again, if unification is not possible due to conflicting constraints (e.g. agreement), the DL tell will fail.

```
find_arg(Functor,Sit,ArgFeat,FinSit) :-
    Arg ?: sign & no(k(inv(dtrs))) in Sit,
    extend_sit(Sit,FinSit),
    Functor :: ArgFeat:Arg in FinSit.
find_arg(Functor,Sit,ArgFeat,FinSit) :-
    new_phrase(Sit,NewSit),
    find_arg(Functor,Sit,ArgFeat,FinSit).
```

The second clause is needed to create a required argument which has not yet been build up. In this case 'new_phrase' is called to create a new potential argument.

For the sentence 'Die schöne Frau sieht sie' we obtain two different parses, since both 'die schoene frau' and 'sie' are ambiguous between nominative and accusative case. The reading according to which 'die schoene frau' is subject is shown in Figure 1 as a phrase structure tree. Some of the corresponding information contained in the DL situation representing this reading is given below:

| | | |
|---|---|---|
| $w_1$ | :: | phon:die & case:nom & start:0 & end:1 |
| $p_8$ | :: | id1 & head_dtr:$p_7$ & comp_dtr1:$w_1$ & start:0 & end:3 |
| $p_9$ | :: | id6 & head_dtr:$w_5$ & start:4 & end:5 |
| $t_{10}$ | :: | trace & tracing:$w_4$ & comp_arg1:$t_{11}$ & comp_arg2:$p_9$ & start:5 & end:5 |
| $t_{11}$ | :: | trace & tracing:$p_8$ & start:4 & end:4 |
| $p_{12}$ | :: | id3 & head_dtr:$t_{10}$ & comp_dtr1:$t_{11}$ & comp_dtr2:$p_9$ & slash:$w_4$ & slash:$p_8$ & start:4 & end:5 |
| $p_{13}$ | :: | id5 & head_dtr:$p_{12}$ & filler_dtr:$w_4$ & slash:$p_8$ & start:3 & end:5 |
| $p_{14}$ | :: | id5 & head_dtr:$p_{13}$ & filler_dtr:$p_8$ & start:0 & end:5 |

In the second parse $t_{11}$ and $p_9$ swap places, i.e. $p_9$ is the 'comp_dtr1' of $p_{12}$ and $t_{11}$ is the 'comp_dtr2'.

The result of the parsing process illustrates the object-centeredness of DL representations. The constituents of the utterance are explicitly modeled and can be used for extracting or specifying further information. Thus we can choose to introduce a feature 'subject' and add the fact '$p_{14}$ :: subject:$p_8$', or we can retrieve all the saturated noun phrases (Phrase ?: np & no(args)). This object-centeredness is useful for disambiguation, for example for anaphora resolution, as illustrated in [Quantz, Schmitz 94].

## 6 Interpretation as Exception Minimization

I will now briefly sketch how the parser presented in the previous section can be extended to perform disambiguation by exception minimization as proposed in [Quantz 93]. In case of ambiguous expressions the parser will return more than one situation. The basic idea of interpretation as exception minimization is to model additional preference rules needed for disambiguation as DL defaults, and to choose the interpretation violating a qualitatively minimal set of defaults.

A Preferential Default Description Logic (PDDL) based on *weigthed defaults* has been developed in [Quantz, Ryan 93]. A weigthed default $\delta$ has the form $c_1 \leadsto_n c_2$, where $c_1$ is called the premise of $\delta$ ($\delta_p$), $c_2$ the conclusions of $\delta$ ($\delta_c$) and $n$ the weight of $\delta$ ($w(\delta)$)—the higher the weight, the more relevant the default. For this PDDL a formally well-behaved preferential entailment relation $^{\mathcal{O}}\!\models_\Sigma$ is presented, which is based on an ordering on DL models $^{\mathcal{O}}\!\sqsubseteq_\Sigma$. The basic idea of this preferential semantics is to compute a *score* for each model by summing up the exceptions to the defaults. Models with lower score are then preferred because they qualitatively minimize the exceptions. It is straightforward to carry the idea of scoring and ordering over from models to situation. To do so, we compute for each situation s and each default $\delta$ the exceptions—those objects for which 'Object ?: $\delta_p$ in s' succeeds and 'Object ?: $\delta_c$ in s' fails.

If there are several possible interpretations for an expression we choose the interpretation given by the situation with the lowest score. (Note that there may be truely ambiguous expressions which yield situations with identical scores.) Thus taking the above example, we might use a preference for topicalization of subjects to prefer the parse shown in Figure 1. This can be achieved by simply introducing a default

$$\text{np \& top:+} \quad \leadsto_5 \quad \text{case:nom}$$

Obviously, this default is a rather weak one and can be overwritten by information stemming from selectional restrictions [Schmitz, Quantz 93].

In principle, it is possible to use preferences stemming from weighted defaults already in the parsing process—situations whose score is higher than a specified threshold are not processed any further. Thus instead of producing all parses in the first step and ordering them in a second step, the parser would then only produce the preferred reading.

## 7 Conclusion

I have presented a DL-based parser for a small HPSG-style fragment of German. Most aspects of the grammar and the parser have been modeled in a highly declarative way. Since the main motivation for my presentation has been to show how an HPSG parser can be implemented in principle by using the inference functionality of a DL system, I have deliberately ignored any efficiency issues. It should be obvious, however, that the parser can be further optimized to increase its performance, for example by integrating chart parsing techniques. We are currently testing the performance of alternative implementations of both the parser and the underlying DL system.

One advantage of using DL as underlying formalism is that in addition to the syntactic information modeled in this paper, semantic and encyclopedic information can be easily integrated into the presented framework. Furthermore, Preferential Default Description Logics can be used to model preference rules as weighted defaults, thereby obtaining interpretation as exception minimization. The parser presented in this paper thus provides the basis for an homogeneous and formally well-founded approach to disambiguation based on Preferential Default Description Logics.